\pdfoutput=1

\documentclass[11pt]{article}

\usepackage[preprint]{acl}

\usepackage{times}
\usepackage{latexsym}
\usepackage{booktabs}
\usepackage{graphicx}
\usepackage{lingmacros}
\usepackage{enumitem}
\usepackage{multirow}
\usepackage{amssymb}

\usepackage{tikz-dependency}
\usepackage{makecell}

\usepackage[T1]{fontenc}

\usepackage[utf8]{inputenc}

\usepackage{microtype}

\usepackage{inconsolata}

\title{Can Language Models Induce Grammatical Knowledge \\ from Indirect Evidence?}

\author{%
Miyu Oba$^{1}$
Yohei Oseki$^{2}$
Akiyo Fukatsu$^{2}$
Akari Haga$^{1}$ \\
\bf
Hiroki Ouchi$^{1}$
Taro Watanabe$^{1}$
Saku Sugawara$^{3}$\\
$^1$Nara Institute of Science and Technology \\
$^2$The University of Tokyo
$^3$National Institute of Informatics \\
\texttt{\{oba.miyu.ol2,haga.akari.ha0,hiroki.ouchi,taro\}@is.naist.jp} \\
\texttt{\{oseki,akiyofukatsu\}@g.ecc.u-tokyo.ac.jp} \\
\texttt{saku@nii.ac.jp}
}

\begin{document}
\maketitle
\begin{abstract}
What kinds of and how much data is necessary for language models to induce grammatical knowledge to judge sentence acceptability?
Recent language models still have much room for improvement in their data efficiency compared to humans.
This paper investigates whether language models efficiently use indirect data (\textit{indirect evidence}), from which they infer sentence acceptability.
In contrast, humans use indirect evidence efficiently, which is considered one of the inductive biases contributing to efficient language acquisition.
To explore this question, we introduce the Wug InDirect Evidence Test (WIDET), a dataset consisting of training instances inserted into the pre-training data and evaluation instances.
We inject synthetic instances with newly coined \textit{wug} words into pretraining data and explore the model's behavior on evaluation data that assesses grammatical acceptability regarding those words.
We prepare the injected instances by varying their levels of indirectness and quantity.
Our experiments surprisingly show that language models do not induce grammatical knowledge even after repeated exposure to instances with the same structure but differing only in lexical items from evaluation instances in certain language phenomena.
Our findings suggest a potential direction for future research: developing models that use latent indirect evidence to induce grammatical knowledge.
\end{abstract}

\section{Introduction}

\begin{figure}[tb]
    \centering
    \includegraphics[width=\linewidth]{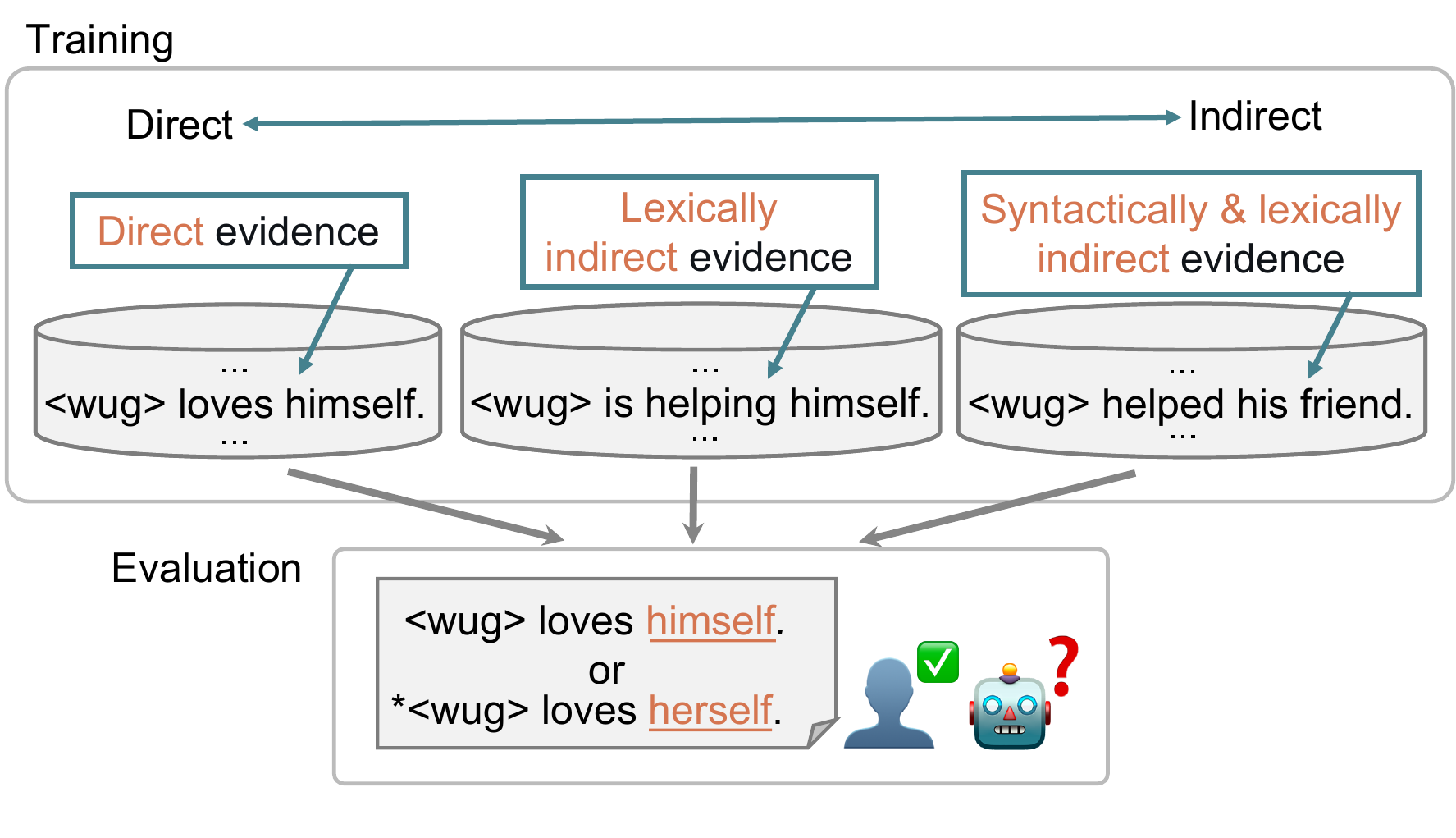}
    \caption{The indirectness of evidence.
    Direct evidence refers to instances identical to previously observed ones.
    Lexically indirect evidence targets the same linguistic knowledge but differs in lexical items.
    Syntactically \& lexically indirect evidence is different in both their syntactical and lexical items.
    }
    \label{fig:overview}
\end{figure}

Recent advances in language models, such as those from the GPT and Llama families~\cite{openai2024gpt4technicalreport,dubey2024llama3herdmodels}, have shown remarkable progress in various tasks.
These models are trained on extremely large datasets, on a scale thousands of times greater than the amount of data children are exposed to in developing grammatical knowledge comparable to that of adults~\citep{warstadt-etal-2023-findings}.
This suggests substantial potential for improving their learning efficiency.

According to \citet{pearl-mis-2016-role}, humans acquire language using \textit{indirect} evidence, in addition to \textit{direct} evidence, which is considered one of the inductive biases contributing to efficient language acquisition.
As illustrated on the left side of Figure~\ref{fig:overview}, when humans encounter the sentence ``<wug> loves himself.'', they can correctly judge the grammatical acceptability between  ``<wug> loves himself.'' and ``*<wug> loves herself.'' 
Such observed sentences are referred to as \textit{direct} evidence. 
Conversely, in the middle and right sides of the figure, we assume that humans are not exposed to such direct evidence.
However, if they observe sentences from which they can make some inference for a correct judgment, such sentences are called \textit{indirect} evidence. 
For example, humans might hypothesize that ``him(self)'' in the sentence ``<wug> is helping himself.'' refers to <wug>, or that the pronoun ``his'' in ``<wug> helped his friend.'' indicates <wug> has a masculine property.

However, it remains still unclear how the degree of indirectness in observed instances affects the number of occurrences required for language models to induce grammatical knowledge.
Previous work has investigated how language models learn grammatical knowledge based on the appearance of items in training data focusing on the word frequency effect~\cite{wei-etal-2021-frequency,yu-etal-2020-word} or generalization to unseen instances ~\cite{patil2024,misra2024,leong2024} through few-shot learning or pretraining on corpora filtered by specific linguistic constructions.
However, those methods face a limitation in identifying ways to enhance the model's learning efficiency.

In this work, we explore the degree of indirectness and the amount of data needed for language models to induce linguistic generalization.
To address this question, we introduce the Wug InDirect Evidence Test (WIDET), a dataset containing additional indirect training and evaluation instances.
We train language models on pretraining data incorporating the indirect training instances.
We then evaluate their linguistic generalization across seven different phenomena, including anaphor agreement, transitivity, and subject-verb agreement.
These phenomena require language models to comprehend diverse properties and multiple parts of speech of specific words to judge their acceptability.
To control the number of observed indirect training instances, we inject synthetic instances with newly coined words into pretraining data.
Following~\citet{berko-wug-1958}, we refer to these words that do not appear in the original vocabulary and data as \textit{wug} words.\footnote{The original \textit{wug} used in~\citet{berko-wug-1958}'s work is not exactly same as our setting to create controlled instances. Details are discussed in Section~\ref{sec:wug_construction}.}
We use various synthetic data as additional indirect training instances, each differing in the degree of lexical and syntactic indirectness as well as the number of observations.

We find that language models generalize linguistic knowledge from training instances identical to correct evaluation instances, through their data efficiency varies across different linguistic phenomena.
This variation is likely influenced by the number of words between the \textit{wug} and the words that act as cues for the model to learn its properties.
We surprisingly observe that the language models do not induce grammatical knowledge in certain phenomena, even in instances that only differ in lexical items.
Syntactically indirect instances rarely induce the model's generalization.

Given that the distances between the \textit{wug} and the cue words to learn its properties might cause inefficiency in the models' learning, we conduct a detailed analysis of indirect instances with complex interference, using anaphor gender agreement as a case study.
We examine whether these instances affect the generalization, considering three factors related to attractors and distance, finding that when the language models are trained on the instances with complex interference, they hit a plateau in learning after sufficient observations.

Those findings from our controlled and comprehensive experiments suggest that, at least in our small-scale settings, language models do not generalize in a human-like manner even from the data with a degree of indirectness that seems intuitively manageable for humans, depending on language phenomena.
Our work contributes insights into language models' capacity to use indirect evidence for learning.
To advance this in future research direction: Implement a model that can use indirect evidence, enabling data-efficient language acquisition comparable to that of humans.\footnote{WIDET is publicly available at \url{https://github.com/nii-cl/widet}.}


\section{Background}
\label{sec:background}
\subsection{Evidence in Language Acquisition}
\label{sec:evidence}
In the field of language acquisition, the information used to learn grammatical knowledge is referred to as \textit{evidence}.
Positive (negative) evidence refers to information in data that indicates what is acceptable (unacceptable) in a language, and it has been argued that humans rely solely on positive evidence to acquire their language~\cite{Chomsky+1993}.
\citet{pearl-mis-2016-role} further distinguishes indirect positive evidence from direct positive evidence. 
Direct positive evidence indicates the information present in the data observed by the learner and used for learning, with the assumption that its usage by speakers guarantees grammaticality (the left side of Figure~\ref{fig:overview}).
Indirect positive evidence, by contrast, refers to information that requires a learner to infer what is grammatical in the language from observed data (the middle and right side of Figure~\ref{fig:overview}).
They argue that, in addition to direct positive evidence, indirect positive evidence potentially plays a significant role in efficient language acquisition.
While the previous literature explores humans' capacity, it is still unclear whether language models induce linguistic generalization from such evidence.

\subsection{Analysis of Language Models in Learning Grammatical Knowledge}
\label{sec:language-model-related-work}

Previous studies have focused on how language models learn grammatical knowledge based on the appearance of target lexical items in training data.
\citet{yu-etal-2020-word} evaluate models' performance on grammatical tasks using minimal pairs including specific target words and few-shot learning on sentences including unseen words.
\citet{wei-etal-2021-frequency} train models on data where the frequency of instances including specific verbs is manipulated to evaluate their generalization to verb inflections.
Recent studies have focused on indirect evidence~\cite{misra2024,leong2024,patil2024}, exploring the availability of indirect evidence in language models by training them from scratch on filtered data.
These data include specific distinctive linguistic phenomena, such as AANN construction~\cite{misra2024} and passivization~\cite{leong2024}, and systematic phenomena from BLiMP~\cite{warstadt-etal-2020-blimp-benchmark}.

\section{Motivations}
\subsection{Experiment Design}
While the previous studies in Section~\ref{sec:language-model-related-work} each offer valuable insights into how language models generalize to unseen instances from various perspectives,
our goal in this work is to explore the impact of the degree of indirectness on data efficiency, with the aim of identifying ways to enhance the model's learning efficiency.
Specifically, we examine how the number of instances required for language models to induce grammatical knowledge changes as the degree of indirectness in the training instances increases.
To archive this, we assume that experiments have to meet the following requirements:

\paragraph{Various Degrees of Indirectness in a Single Linguistics Phenomenon} To investigate the impact of the degree of indirectness on the number of observations needed for grammar acquisition, we employ two graduated types of indirectness, lexical and syntactic, in addition to direct evidence.
Most prior research focuses on a single degree of indirectness for a given linguistic phenomenon.

\paragraph{Various Number of Observations}
Given our aim for data efficiency, we need to quantify how much the required amount of data for language models to induce grammatical knowledge increases due to indirectness.
We employ six different observation counts, ranging from 0 to 100.
Previous studies focusing on indirect evidence are limited in their ability to quantify changes in the number of observations required, as they do not take into account the frequency effect.

\paragraph{Various Linguistics Phenomena} 
We explore whether the two aspects mentioned above occur universally across linguistic phenomena or are specific to certain phenomena.
We employ seven types of linguistics phenomena, each with referent targets consisting of several different parts of speech.
Most of the previous work, except for \citet{patil2024}, focuses on one or two phenomena.

\paragraph{Inserting Sentences Containing Words that do not Appear in Pretraining Data} Considering phenomena like anaphor agreement, to judge the acceptability of a sentence, language models are expected to understand the properties (e.g., number) of the referent in the sentence.
To count the number of observations for language models to induce grammatical knowledge, we need to concisely count how many times the language models encounter a sentence containing the referent before they understand the properties of the referent.
For conventional approaches to ablate or certain lexical items existing in corpora, the (sub)word of the target referent may appear in the sentence other than the removed one, making it difficult to count the observations accurately.
To concisely control the number of observations of the referent, we employ the sentences containing the words that have not appeared in pretraining corpora.

\begin{table*}[t]
    \resizebox{\linewidth}{!}{
    \small
    \begin{tabular}{ccll} 
    \toprule
    Phenomenon & \textit{Evd} & Training instance & Evaluation instance \\ 
    \midrule
    \multirow{3}{*}{\makecell{Anaphor gender agreement\\(\textsc{Ana.gen.agr})}} & DE & <wug\#n> has devoted herself &  \multirow{2}{*}{<wug\#n> has devoted herself} \\
     & LexIE & <wug\#n> is painting herself & \multirow{2}{*}{*<wug\#n> has devoted himself} \\
     & SynIE & <wug\#n> judges her work & \\
    \cmidrule(lr){1-1} \cmidrule(lr){2-2} \cmidrule(lr){3-3} \cmidrule(lr){4-4}
    \multirow{3}{*}{\makecell{Anaphor number agreement\\(\textsc{Ana.num.agr})}} & DE & the <wug\#n> didn't see themselves & \multirow{2}{*}{the <wug\#n> didn't see themselves} \\
     & LexIE & the <wug\#n> can reward themselves & \multirow{2}{*}{*the <wug\#n> didn't see itself} \\
     & SynIE & the <wug\#n> loved their toy & \\
     \cmidrule(lr){1-1} \cmidrule(lr){2-2} \cmidrule(lr){3-3} \cmidrule(lr){4-4}
    \multirow{3}{*}{\makecell{Transitive\\(\textsc{Trans.})}} & DE & some trees <wug\#n>ed the car & \multirow{2}{*}{some trees <wug\#n>ed the car} \\
     & LexIE & no street can <wug\#n> the city & \multirow{2}{*}{*some trees <wug\#n>ed}\\
     & SynIE & every lion hunts what no prey can <wug\#n> & \\
     \cmidrule(lr){1-1} \cmidrule(lr){2-2} \cmidrule(lr){3-3} \cmidrule(lr){4-4}
    \multirow{3}{*}{\makecell{Intransitive\\(\textsc{Intrans.})}} & DE & many rivers should <wug\#n> & \multirow{2}{*}{many rivers should <wug\#n>} \\
     & LexIE & each ethic might <wug\#n> & \multirow{2}{*}{*many rivers should <wug\#n> dogs}\\
     & SynIE & a man corrects that the answer will not <wug\#n> & \\
     \cmidrule(lr){1-1} \cmidrule(lr){2-2} \cmidrule(lr){3-3} \cmidrule(lr){4-4}
    \multirow{3}{*}{\makecell{Determiner-Noun agreement\\(\textsc{D-N agr})}} & DE & the senators use this <wug\#n> & \multirow{2}{*}{the senators use this <wug\#n>} \\
     & LexIE & a window will open this <wug\#n> & \multirow{2}{*}{*the senators use these <wug\#n>} \\
     & SynIE & the <wug\#n> sells the house &  \\
     \cmidrule(lr){1-1} \cmidrule(lr){2-2} \cmidrule(lr){3-3} \cmidrule(lr){4-4}
    \multirow{3}{*}{\makecell{Subject-Verb agreement (V)\\(\textsc{S-V agr (V)})}} & DE & the <wug\#n> are leaving any traces & \multirow{2}{*}{the <wug\#n> are leaving any traces} \\
     & LexIE & the <wug\#n> climb few ladders & \multirow{2}{*}{*the <wug\#n> is leaving any traces} \\
     & SynIE & each key can open those <wug\#n> & \\
     \cmidrule(lr){1-1} \cmidrule(lr){2-2} \cmidrule(lr){3-3} \cmidrule(lr){4-4}
    \multirow{3}{*}{\makecell{Subject-Verb agreement (S)\\(\textsc{S-V agr (S)})}} & DE & the book <wug\#n> a shelf & \multirow{2}{*}{the book <wug\#n> a shelf} \\
     & LexIE & every chocolate <wug\#n> several bars & \multirow{2}{*}{*the books <wug\#n> a shelf} \\
     & SynIE & the deer that trails the head <wug\#n> a herd & \\
    
    \bottomrule
    \end{tabular}
    }
    \caption{Linguistic phenomena and instances. The sentences starting with * are ungrammatical.}
    \label{tab:phenomena}
\end{table*}

\begin{table}[t]
    \centering
    \resizebox{\linewidth}{!}{
    \begin{tabular}{lccccc} 
    \toprule
    Phenomenon &  POS & Gen. & Num. & (In)Transitive & Long agr\\
    \midrule
    \textsc{Ana.gen.agr}. & noun & \checkmark & -- & -- & \checkmark \\
    \textsc{Ana.num.agr} & noun & -- & \checkmark & -- & \checkmark \\
    \textsc{Trans.} & verb & -- & -- & \checkmark & -- \\
    \textsc{Intrans.} & verb & -- & -- & \checkmark & -- \\
    \textsc{D-N agr} & adj & -- & \checkmark & -- & -- \\
    \textsc{S-V agr (v)}  & verb & -- & \checkmark & -- & -- \\
    \textsc{S-V agr (s)} & noun & -- & \checkmark & -- & -- \\
    \bottomrule
    \end{tabular}
    }
    \caption{Properties to judge evaluation data. POS denotes part-of-speech. Gen./Num. denotes gender/number. Long agr. is whether a long agreement is required.}
    \label{tab:properties}
\end{table}

\subsection{Inserting Instances with Newly Coined Words}
\label{sec:motivation}
We employ newly coined words (\textit{wugs}) to introduce additional instances including words that do not appear in pretraining data.
The advantages include:

\begin{itemize}[leftmargin=1em]
    \item Handling the occurrences of target lexical items may not eliminate their influence from the pretraining corpus.
    To fully negate the effect of a lexical item, all variants sharing the same stem or subword would need to be removed, which is complex and risks significantly distorting the natural corpus distribution.
    \item When automatically generating \textit{wugs}, we can adequately control their frequency and evidence strength, including their tokenization. 
    Since our aim here is to control the minimal information observable by the model, synthetic data allows for the elimination of noises.
    \item Our approach is a form of data augmentation, that does not require any modification of lexical items or sentences in the corpora. Hence, it can be easily applied to other corpora and models.
\end{itemize}

While using artificial languages in analyzing language models is tackled by previous work \citep{white-cotterell-2021-examining,ri-tsuruoka-2022-pretraining}, our approach is different in that we use artificial instances only at the token level by introducing a word \textit{wug} to insert them into a natural corpus. 

\section{Wug InDirect Evidence Test (WIDET)}
This section describes how we construct additional training and evaluation instances, which comprise our dataset, WIDET.
Following targeted syntactic evaluation \citep{linzen-etal-2016-assessing,marvin-linzen-2018-targeted,warstadt-etal-2020-blimp-benchmark}, we employ minimal pair paradigm where pairs of sentences minimally differ in target words.
The examples of instances are listed in Table~\ref{tab:phenomena}.

\subsection{Linguistic Phenomena}
We employ the seven different linguistic phenomena listed in Table~\ref{tab:phenomena}, which we selected from the benchmark BLiMP~\cite{warstadt-etal-2020-blimp-benchmark}\footnote{Appendix~\ref{sec:from_blimp} details the specific phenomena referenced from BLiMP in this work.}.
As shown in Table~\ref{tab:properties}, the phenomena vary in their properties, so that we can analyze models' behavior from diverse perspectives.
Since our selection criteria are based on whether understanding the properties of a single word is sufficient to judge the linguistic phenomena correctly, we can only cover limited linguistic phenomena.
We anticipate phenomena related to island effects, for instance, to be beyond this scope. 

\subsection{Newly Coined Word \textit{Wug}}
We employ the tag <wug\#n> as a newly coined word to conduct controlled experiments using words that never appeared in the pretraining corpus.
This approach does not entirely align with the policy in~\citet{berko-wug-1958}, which employed words like \textit{wug} and \textit{wuz} that are newly coined but phonologically natural in the target language by using actual subwords. One concerning issue with ~\citet{berko-wug-1958}'s policy is that the actual subwords can provide models with clues for correct grammatical judgment, for example, by their occurrence in specific positions.
While using actual subwords could help models access grammatical knowledge needed for accurate judgment, it complicates evaluating the models' true ability to learn from indirect evidence.
To avoid its possible effects, we instead use the artificial tag <wug\#n>.
We analyze the differences between the conditions using the tag and the original \textit{wug} in Section~\ref{sec:wug_construction}.

\subsection{Indirectness of Additional Training Instances}
\label{sec:additional_training_instances}
We define the following three degrees of indirectness (DE, LexIE, and SynIE).
The difficulty increases in the order of DE, LexIE, and SynIE: 

\paragraph{Direct Evidence (DE)} An instance identical to the correct evaluation instances.
We assume that the properties of \textit{wug} in an evaluation instance are learned by referencing the training instance that shares the same syntactical and lexical items as the evaluation instance.
\paragraph{Lexically Indirect Evidence (LexIE)} An instance that conveys the same syntactic structure as the evaluation instance but uses different lexical items.
We assume that the properties of \textit{wug} in an evaluation instance are learned by referencing training instances with the same usage but different lexical items from those in the evaluation instance.
\paragraph{Syntactically Indirect Evidence (SynIE)} An instance that reveals the target linguistic feature with different syntactic and lexical items from evaluation instances.
The properties of \textit{wug} in an evaluation instance are learned by referencing the training instance with different syntactic and lexical items from those in the evaluation instance.

\subsection{Training and Evaluation Template}
We prepare 200 template pairs for each linguistic phenomenon.
Each template has three different sets of tags, resulting in $200 \times 3 = 600$ pairs.

We anticipate that quantifiers and determiners can influence linguistic generalization, making it unclear whether language models rely on the properties of verbs and reflexive pronouns, quantifiers, and determiners, or other factors as clues for judgment, while previous studies have paid limited attention to this~\cite{patil2024}.
To mitigate such effects, for number agreement, we added <wug\#n> without any suffixes to these sentences, expecting the models to infer that <wug\#n> is an inflected form based on the sentence structure in which they are embedded.
We explore their effects in the model's generalization in Section~\ref{sec:wug_construction}.
For the noun subject of \textsc{S-V agr (V)} and \textsc{Ana.num.agr}, we avoid any quantifiers and determiners other than ``the''.
Due to the same reason, for the verb in \textsc{S-V agr (S)}, we only employ the present tense and do not employ any auxiliary verbs and tense suffixes.
We ensured that <wug\#n> was used the same word (i.e., the tag with the same id) in a pair, both grammatical and ungrammatical sentences because we want the same occurrence of the \textit{wug} in the training data.

\subsection{Data Generation with LLM}
To create varied degrees of and balanced corpus, we use GPT-4 Turbo in OpenAI API to generate the training and evaluation templates.
To generate balanced training instances with different properties, we generate them separately based on concerning properties, (e.g., feminine and masculine pronouns have the same percentage in \textsc{Ana.gen.agr}.).
We prompt the GPT-4 to generate balanced, diverse, and duplication sentences.
We generate evaluation instances and training instances for indirect evidence (LexIE, SynIE) with three different prompts.
Subsequently, we get DE by extracting the correct sentence in generated evaluation instances.
We generate the sentences with placeholders [WUG] and we replace [WUG] with the tag <wug\#n>, where the index number $n$ distinguishes the coined words (e.g., <wug\#124>).
The example of prompts and detailed procedures are shown in Appendix~\ref{sec:all_other_prompt_examples}.

\section{Experiments and Results}
\label{sec:experiments}

\subsection{Settings}

\begin{figure*}[tbp]
\centering
\includegraphics[width=\linewidth]{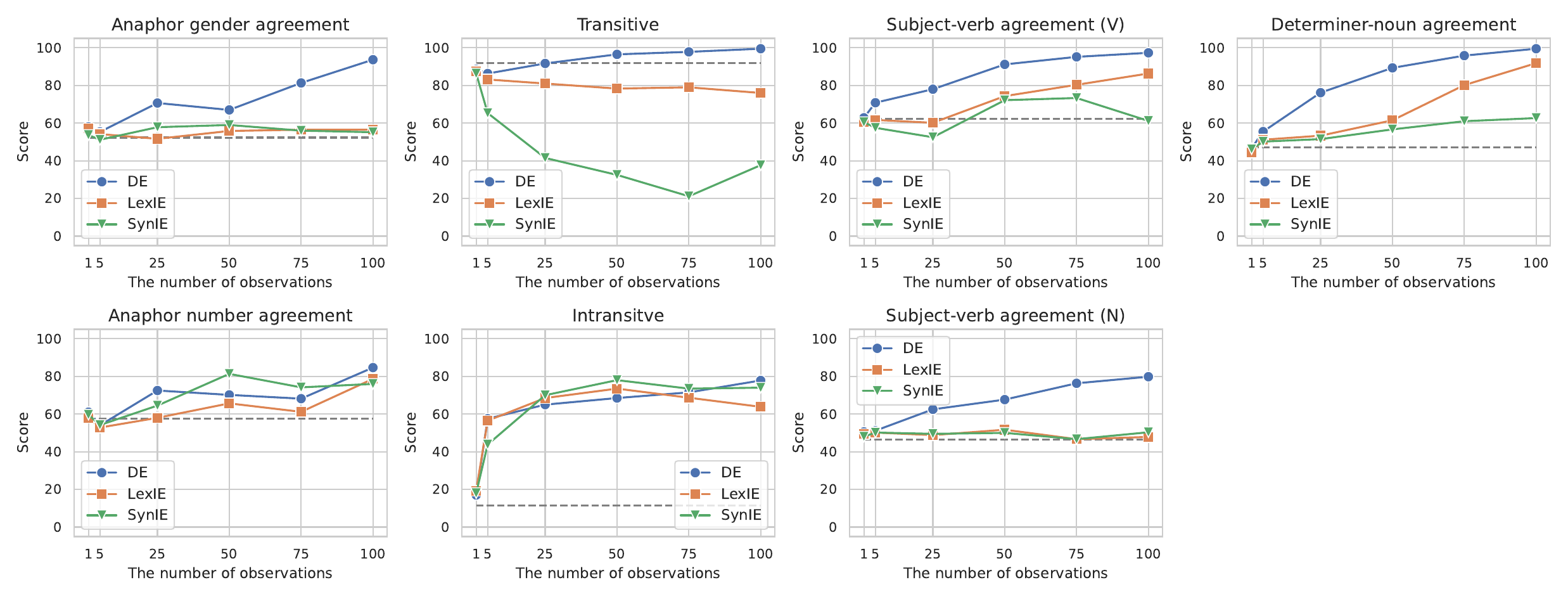}
\caption{The results (accuracy; \%) of experiments for language phenomena and evidence. The gray dot lines indicate the model's scores trained on pretraining data without any additional instances (n=0).}
\label{fig:main_results}
\end{figure*}

\paragraph{Pretraining Data}
We randomly sample 675k sentences (16M words) from English Wikipedia articles and use them as pretraining data.\footnote{Retrieved from \url{https://github.com/phueb/BabyBERTa}.}
We inject the additional training instances into the data.
The detailed preprocessing steps and additionally injected training instances are described in Appendix~\ref{sec:detailed_data_generation}.
We shuffle and deduplicate sentences and remove ones containing fewer than two words. The data is then lowercase, and periods are removed from the sentences.

\paragraph{Frequency of Additional Instances}
We compare the language models trained on the pretraining data injected indirect instances that appear $n$ times ($n=0, 1, 5, 25, 50, 75, 100$) for each instance.

\paragraph{Models}
We use BabyBERTa~\citep{huebner-etal-2021-babyberta}, which is a minimal variant of RoBERTa~\citep{liu2019roberta}.
We modify some hyperparameters due to the pretraining data size.
More detailed information is shown in Table~\ref{tab:hypara}.
We train the tokenizer from scratch on the pretraining data, adding the tags to the vocabulary so that the tokenizer treats each tag as one token.

\paragraph{Evaluation Metrics}
We use the accuracy of selecting the correct sentence as our evaluation metric.
We employ pseudo-likelihood~\cite{salazar-etal-2020-masked}\footnote{We use the source code in \url{https://github.com/babylm/evaluation-pipeline-2023}.} normalized by token length because we use evaluation sentences containing the sentence pair each of which has different token lengths.~\footnote{Normalization by token length may still result in token-biases~\cite{ueda-etal-2024-token-length}.}

\subsection{Results}
\label{sec:results}

We review the main results by answering our research questions: (i) What degree of and how much data do language models need to acquire grammatical knowledge to judge the acceptability of a sentence?
(ii) Are observations showing similar trends in broader categories of linguistic phenomena?
The results are shown in Figure~\ref{fig:main_results}.

\paragraph{Direct Evidence}
As for DE, increasing the number of observations generally contributed to linguistic generalization in language models.
However, the extent of improvement varied across different linguistic phenomena.
In \textsc{Ana.gen.agr} and \textsc{Ana.num.agr}, the score increased more gradually, particularly between 25 and 75 occurrences, compared to the other agreement phenomena.
This difference might be due to anaphor agreement, which often involves a longer distance between the target words and the words with properties necessary for correct judgment.
We thoroughly examine the effects of distance and attractors in Section~\ref{sec:longer_agreement}.

\paragraph{Lexically Indirect Evidence}
In about a half of the phenomena, \textsc{D-N agr}, \textsc{S-V agr (V)}, \textsc{Ana.num.agr}, and \textsc{Intransitive}, LexIE induces generalization more slowly but steadily than DE.
However, in the remaining half of the phenomena, the language models do not acquire the grammatical knowledge necessary to correctly judge acceptability.
This result is surprising because LexIE differs only in lexical items from a correct sentence in the evaluation and shares the same syntactical structure. 
This trend cannot be explained by the properties of Table~\ref{tab:properties}.
\paragraph{Syntactically Indirect Evidence}
In most phenomena, the models fail to induce SynIE generalization;  the increase in the number of observations did not improve generalization but merely extended learning time.
In \textsc{Transitive}, the accuracy of SynIE drastically decreases inversely with the number of observations. 
This intriguing phenomenon is likely due to the heuristics of the language model.
The final word in the training instances (see Table~\ref{tab:phenomena}) is the <wug\#n>, whereas it is an actual direct object noun in the correct evaluation sentences. 
This suggests that the language model might exhibit linear generalization~\cite{mueller-etal-2022-coloring,mccoy-etal-2020-syntax}, which differs from the human-like hierarchical generalization.
The model seems to judge correctness based on whether certain words follow the <wug\#n>, even though the \textit{wug} should be recognized as a transitive verb because the relative pronoun ``what'' is its object.
This implies that instances requiring complex hierarchical inference may hinder generalization.

\paragraph{Overall}
Our findings mainly suggest that language models do not sufficiently induce linguistics generalization from indirect positive evidence, especially SynIE, while they induce it from direct evidence.
\citet{wei-etal-2021-frequency} find that their results support the Reduce Error Hypothesis~\cite{AMBRIDGE_KIDD_ROWLAND_THEAKSTON_2015}, where high-frequency words are learned better.
The results in our work also support the hypothesis in DE, but in LexIE and SynIE, not all linguistic phenomena support it.

\section{Analysis with More Indirect Instances}
\label{sec:longer_agreement}

\begin{table}[tbp]
    \centering
    \resizebox{\linewidth}{!}{
    \small
    \begin{tabular}{p{1.1cm}p{0.6cm}p{4.9cm}} 
    \toprule
    Interf. & Evd. & Training instance \\ 
    \midrule
    \multirow{5}{*}{\makecell{Attractor\\type\\(AT)}} & DE & <w> loves herself \\
    \cmidrule[0.03em](lr){2-2} \cmidrule[0.03em](r){3-3}
    & AT0 & <w> \textbf{helping the child} loves herself \\
    \cmidrule[0.03em](lr){2-2} \cmidrule[0.03em](r){3-3}
    & AT1 & <w> \textbf{helping the man} loves herself \\
    \cmidrule[0.03em](lr){2-2} \cmidrule[0.03em](r){3-3}
    & AT2 & <w> \textbf{helping him} loves herself \\
    \midrule
    \multirow{10}{*}{\makecell{Attractor\\number\\(AN)}} & DE & <w> loves herself \\
    \cmidrule[0.03em](lr){2-2} \cmidrule[0.03em](r){3-3}
    & AT1 & <w> \textbf{helping the man} loves herself\\
    \cmidrule[0.03em](lr){2-2} \cmidrule[0.03em](r){3-3}
    & AN0 & <w> \textbf{helping the man to see the dad} loves herself \\
    \cmidrule[0.03em](lr){2-2} \cmidrule[0.03em](r){3-3}
    & AN1 & <w> \textbf{helping the man for the king to see the dad} loves herself \\
    \cmidrule[0.03em](lr){2-2} \cmidrule[0.03em](r){3-3}
    & AN2 & <w> \textbf{helping the man for the son of the king to see the dad} loves herself \\
    \midrule
    \multirow{8}{*}{\makecell{Distance\\(DT)}} & DE & <w> loves herself \\
    \cmidrule[0.03em](lr){2-2} \cmidrule[0.03em](r){3-3}
    & AT0 & <w> \textbf{helping the child} loves herself \\
    \cmidrule[0.03em](lr){2-2} \cmidrule[0.03em](r){3-3}
    & DT0 & <w> \textbf{who helps the child} loves herself \\
    \cmidrule[0.03em](lr){2-2} \cmidrule[0.03em](r){3-3}
    & DT1 & <w> \textbf{whose cat helps the child} loves herself \\
    \cmidrule[0.03em](lr){2-2} \cmidrule[0.03em](r){3-3}
    & DT2 & <w> \textbf{whose cat helps the child who finds the teachers} loves herself \\
    \bottomrule
    \end{tabular}
    }
    \caption{Interference types and training instances used in the analysis. <w> corresponds to <wug\#n>.}
    \label{tab:analysis_phenomena}
\end{table}

In Section~\ref{sec:experiments}, DE induced the model's linguistic generalization but its data efficiency varies by linguistic phenomena.
For anaphor agreement, the models' learning is more apt to reach a plateau in 25 -- 75 observations compared to other phenomena (See the figure for anaphor agreement in Figure~\ref{fig:main_results}).
This stagnation might be caused by the longer distance between the \textit{wug} and the reflexives, whereas the relevant items are adjacent to each other in other phenomena such as \textsc{Transitive}.
To corroborate this negative effect of long distance on learning, we employ more indirect agreement instances to investigate whether the long distance hinders linguistic generalization on \textsc{Ana.gen.agr} in language models. 

The difficulty of long-distance agreement is caused by attractors and distance~\citep{linzen-etal-2016-assessing}.
Agreement attractors indicate the intervening words that distract the learner from judging the correct agreement~\citep{giulianelli-etal-2018-hood}.
When language models judge the gender agreement, they would check if the word ``<wug\#n>'' corresponds to the gender of the reflexive.
\textit{Distance} refers to the number of the words intervening between the antecedent ``<wug\#n>'' and ``herself''.
\textit{Attractor} indicates the competing words (e.g., ``man'' in the case of AT1 in Table~\ref{tab:analysis_phenomena}) that distract learners from judging the agreement.

The language models' grammatical knowledge concerning long-distance dependencies has been investigated in previous studies~\citep{giulianelli-etal-2018-hood,li-etal-2023-assessing}, and these studies argue that the models can indeed acquire the knowledge of long-distance agreement. However, the overall results on anaphor agreement in this study suggest that further investigation is required to reveal the relationship between models' performance and the distance of items relevant to correct judgment. For this purpose, we conduct a fine-grained analysis using synthetic sentences varying the distance between~\textit{wugs} and reflexive pronouns.

\subsection{Target Phenomena}
We compare the models trained on the corpus with additional instances of anaphor gender agreement, from the perspective of the attractor type, number, and distance as below.
Table~\ref{tab:analysis_phenomena} lists all kinds of training instances compared in this analysis.

To create the instances, we use GPT-4 to generate nouns differing in gender and number and sample the designated number of items from these generated items. 
For feminine and masculine nouns, we collect 100 nouns each. From the generated items, we first select 25 nouns for each gender. Then, we create both the singular and plural forms of the selected words and double them to create minimal pairs.
The prompt is shown in Appendix~\ref{sec:all_other_prompt_examples}.
Additionally, we also collect 100 neutral nouns such as teacher and child. 
The verb that we newly employ is collected from LexIE in \textsc{Ana.gen.agr} to avoid duplication.

\paragraph{Attractor Type (AT)}
We investigate whether attractors downgrade the linguistic generalization in \textsc{Ana.gen.agr} and how their distract strength affects the models' acquisition of anaphor agreement.
DE indicates the direct instances examined in Section~\ref{sec:experiments}, which does not have any attractors and works as a baseline here.
AT0 includes neutral common nouns, while AT1 employs common opposite-gender nouns, and AT2 uses opposite-gender pronouns.
We assume that the magnitude of attractors' interference follows the order AT0 < AT1 < AT2, given that the more similar their properties are to reflexives, the more distracting they will be.

\paragraph{Attractor Number (AN)}
We examine whether the number of attractors affects the model's acquisition.
We use the gender common nouns as attractors.
DE works as a baseline because it has no attractors. 
We expect that the more attractors there are, the more difficult it is to generalize correctly.

\paragraph{Distance (DT)}
We analyze the effect of distance on the model's acquisition.
We assume that the more distance intervening between \textit{wug} and reflexive, the more difficult it is to judge sentence acceptability.
We use neutral nouns there to explore the effect of the number of words genuinely.

\subsection{Results}
\begin{figure}[tbp]
    \centering
    \includegraphics[width=\linewidth]{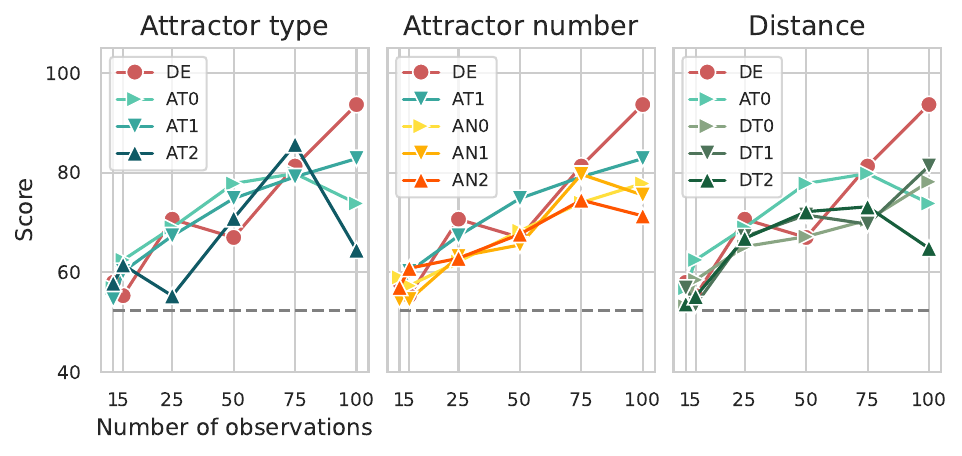}
    \caption{Models' scores for more indirect instances.}
    \label{fig:analysis_results}
\end{figure}

As shown in Figure~\ref{fig:analysis_results},
After 100 observations in all viewpoints, SynIE, with the shortest distance and no attractors, got the highest scores, while in midway observations this tendency does not happen.
The most difficult instances in each interference lead to the language model's lowest score, after their 100 observations.
AT2, including an opposed pronoun as an attractor, particularly shows unstable generalization.
We initially expected that instances with longer distances and more attractors would interfere more strongly with the models' generalization.
However, this tendency was not observed in the experiment.
To the question of whether the instances with long-distance agreement induce linguistic generalization, these results answer that with the larger number of observations, the model's generalization relatively hits a plateau.

\section{Discussion}
\subsection{Considering \textit{Wug} Creation}
\label{sec:wug_construction}
In this work, we use newly coined words that do not appear in the original vocabulary, following ~\citet{berko-wug-1958}.
Still, our used \textit{wug} has some gap from the original one.
In the original \textit{wug} test, they use the words that do not exist in the language but conform to the phonological rule in the language.
In contrast, we use the tag <wug\#n> as \textit{wug} in those experiments.
Since the original \textit{wug} is more phonologically natural, and the subwords are in the existing vocabulary, the original setting is closer to the environment of human language acquisition.
On the other hand, to conduct controlled experiments on the number of instances that the model observed, the setting might not be suitable because this is far from the settings where a certain word is never encountered.
We used the tag <wug\#n>.
In this section, we compare our method (\textit{tag} method) and the original method (\textit{wug} method) to explore the difference in their impact on the model's linguistic generalization.

\paragraph{\textit{Wug} Generation}
We create \textit{wug} using pseudoword generator Wuggy.\footnote{\url{https://github.com/WuggyCode/wuggy}.} and choose 1.2k nouns from sample data taken from the one billion-word Corpus of Contemporary American English (COCA).\footnote{Downloaded from \url{https://www.wordfrequency.info/samples/words_219k.txt}.}
To create \textit{wug}-like words, we use the nouns to output four pseudo words for one noun and randomly select one pseudo noun.
We prepare $200 \times 3 = 600$ pseudo words, each 200 of which are used separately (\textit{wug\_v1--wug\_v3}) because we expect that different \textit{wug}s have different subwords and they can show different results.
\footnote{On the other hand, for \textit{tag} and \textit{tag w/ morph.}, we show the results of only one model, because the different \textit{tag}s <wug\#n> have the same parameters and they actually show the same results.}
We use those pseudo nouns instead of the tag in the same way as in the previous experiments.

\paragraph{Settings}
We target three phenomena, \textsc{ana.num.agr}, \textsc{d-n agr}, and \textsc{s-v agr~(v)}, the \textit{wug} of which is considered as common nouns.
No inflectional morphemes are added to plural common nouns in the \textit{tag} method while the morphemes are added to plural common nouns in the \textit{wug} method.
For ablation, we prepare the tag with inflectional morphemes (\textit{tag w/ morph.} method), which employs the tag <wug\#n> same as the \textit{tag} method but uses inflectional morphemes same as the \textit{wug} method.
We compare the models trained on the pretraining data with the \textit{tag} method, the \textit{wug} methods, and \textit{tag w/ morph.} method.
Other settings are the same as Section~\ref{sec:experiments}.

\begin{table}[tbp]
    \centering
    \small
    \resizebox{\linewidth}{!}{
    \begin{tabular}{llccc} 
    \toprule
    \multirow{2}{*}{N} & \multirow{2}{*}{\textit{wug} method} & \multicolumn{3}{c}{Phenomenon} \\
     &  & \textsc{ana. num. agr} & \textsc{d-n agr} & \textsc{s-v agr (v)}  \\
    \midrule
    \multirow{5}{*}{0} & \textit{tag} & 57.5 & 47.0 & 62.2 \\
    & \textit{tag w/ morph.} & 59.0 & 80.5 & 83.3\\ 
    & \textit{wug\_v1} & 81.3 & 89.5 & 86.7 \\
    & \textit{wug\_v2} & 81.2 & 91.2 & 86.0 \\
    & \textit{wug\_v3} & 81.5 & 88.7 & 85.0\\
    \midrule
    \multirow{5}{*}{25} & \textit{tag} & 72.5 & 76.2 & 78.0 \\
    & \textit{tag w/ morph.} & 94.0 & 99.5 & 91.3\\
    & \textit{wug\_v1} & 92.3 & 87.7 & 90.2 \\
    & \textit{wug\_v2} & 90.5 & 87.7 & 88.5 \\
    & \textit{wug\_v3} & 90.5 & 87.5 & 86.5 \\
    
    \bottomrule
    \end{tabular}
    }
    \caption{Scores calculated by the models trained on the pretraining data with indirect instances of different \textit{wug} creation methods. \textit{N} is the number of observations.}
    \label{tab:wug_results}
\end{table}

\paragraph{Results}
Table~\ref{tab:wug_results} shows the scores of the \textit{tag}, \textit{tag w/ morph.}, and three sets of \textit{wug}.
In the \textit{wug} and \textit{tag w/ morph.}, the language models correctly judge the acceptability of sentences, mostly more than 80--90\%, surprisingly with the data that includes zero additional instances.
This result is probably because language models determine whether a word is singular or plural, based on whether an inflection morpheme ``s'' follows it, even if the word is novel.
This occurs with both novel words and novel subword combinations, but the impact is greater with the latter, comparing the two methods.
In addition, despite our expectation that different subword combinations show different results, we observed no large score variances among the three vocabulary sets except for 25 times in \textsc{ana.num.agr}.
From those results, we found a trade-off between the settings plausible for human language acquisition and strictly controlled settings.
We prioritized the latter in this work, but the direction to the former is also a good setting depending on the research questions.

\subsection{Zero Observations of \textit{Wug}}
While a tag <wug\#n> is added to the vocabulary, its parameters in language models are randomly initialized.
If the language models never encounter sentences containing this tag during training, its parameters still remain in their initialized state, which may lead to varying results in language models depending on factors such as the initializer’s standard deviation (std) and the random seed used.
To verify this effect, we compare the language model using the default std of the initializer for all weight matrices (std $=0.02$) to that with one-tenth std (std $=0.002$), using three kinds of seeds.
Table~\ref{tab:seeds} shows that the deviation of scores is smaller in the model using one-tenth std for initializer compared to the model using the default std.
This finding implies that a smaller std can enhance the stability of the results.
However, an excessively small std may risk negatively affecting the training process.
Hence, we employ default std in the current work.

\begin{table}[tbp]
    \centering
    \small
    \begin{tabular}{@{}lcc@{}}
    \toprule
    Phenomenon & Std & Score \\
    \midrule
    \multirow{2}{*}{\textsc{Ana.gen.agr}} & 0.02 & $51.3\pm0.95$ \\
    & 0.002 & $55.5\pm1.73$ \\
    \cmidrule(r){1-1} \cmidrule(lr){2-2} \cmidrule(l){3-3}
    \multirow{2}{*}{\textsc{Ana.num.agr}} & 0.02 & $59.7\pm2.44$ \\
    & 0.002 & $64.4\pm2.84$ \\
    \cmidrule(r){1-1} \cmidrule(lr){2-2} \cmidrule(l){3-3}
    \multirow{2}{*}{\textsc{Transitive}} & 0.02 & $90.2\pm1.57$ \\
    & 0.002 & $90.0\pm1.15$ \\
    \cmidrule(r){1-1} \cmidrule(lr){2-2} \cmidrule(l){3-3}
    \multirow{2}{*}{\textsc{Intransitive}} & 0.02 & $12.7\pm1.53$  \\
    & 0.002 & $12.0\pm0.60$ \\
    \cmidrule(r){1-1} \cmidrule(lr){2-2} \cmidrule(l){3-3}
    \multirow{2}{*}{\textsc{D-N agr}} & 0.02 & $47.4\pm1.39$  \\
    & 0.002 & $48.9\pm1.68$ \\
    \cmidrule(r){1-1} \cmidrule(lr){2-2} \cmidrule(l){3-3}
    \multirow{2}{*}{\textsc{S-V agr (V)}} & 0.02 & $56.4\pm5.23$ \\
    & 0.002 & $54.7\pm1.78$ \\
    \cmidrule(r){1-1} \cmidrule(lr){2-2} \cmidrule(l){3-3}
    \multirow{2}{*}{\textsc{S-V agr (S)}} & 0.02 & $49.1\pm2.98$ \\
    & 0.002 & $49.4\pm1.19$ \\ 
    \bottomrule
    \end{tabular}
    \caption{Scores (mean$\pm$std) of language models with different seeds and standard deviation of the initializers.}
    \label{tab:seeds}
\end{table}

\section{Conclusion}
We investigate the degree of indirectness and the amount of data required to induce human-like linguistic generalization in language models.
We found that language models do not induce human-like linguistic generalization even with a degree of indirectness that seems intuitively manageable for humans, depending on language phenomena.
This limitation indicates a direction for future studies: implementing a model that can use indirect evidence, which will lead to data-efficient language acquisition comparable to that of humans.

\newpage
\section*{Limitations}

We recognize the following limitations in this study:

\paragraph{Linguistic Knowledge by Function Words}
We generate synthetic instances only for linguistic phenomena concerning content words such as nouns and verbs.
We avoid generating new function words (e.g., new \textit{wh}-word as a relative pronoun).

\paragraph{Nonce Sentence}
We have not dug into the difference between natural sentences and nonce sentences~\cite{gulordava-etal-2018-colorless,wei-etal-2021-frequency} that are grammatical but completely meaningless because we create additional training and evaluation instances with LLM, which tends to generate naturally plausible sentences.
Nonce sentences are less plausible in human language acquisition but exclude semantic selectional-preferences cues~\cite{gulordava-etal-2018-colorless,goldberg2019assessing}. 
According to Section~\ref{sec:wug_construction}, there can be a trade-off between training language models in experimental settings that closely resemble natural human language acquisition and those that are strictly controlled.
Future work can determine whether nonce sentences with indirect evidence differently affect linguistic generalization in language models.

\paragraph{Limited Model Size and Pretraining Data}
We use a small-scale language model and pretraining data in this work because we aim to find the differences from human inductive biases as much as possible.
It is uncertain that the same trends as our work will appear in models of any size.
Whether scaling laws apply to indirect data in accelerating model generalization would be an interesting future work.

\section*{Ethics Statement}
There might be a possibility that the texts we used (Wikipedia) and the sentences generated by large language models are socially biased, despite their popular use in the NLP community.

\section*{Acknowledgments}
We would like to express our gratitude to the anonymous reviewers who provided many insightful comments that have improved our paper.
This work was supported by JSPS KAKENHI Grant Numbers JP21H05054, 22K17954, and 24KJ1700, and JST PRESTO Grant Numbers JPMJPR21C2 and JPMJPR20C4.

\bibliography{anthology,custom}


\begin{figure*}[htbp]
\centering
\includegraphics[width=\linewidth]{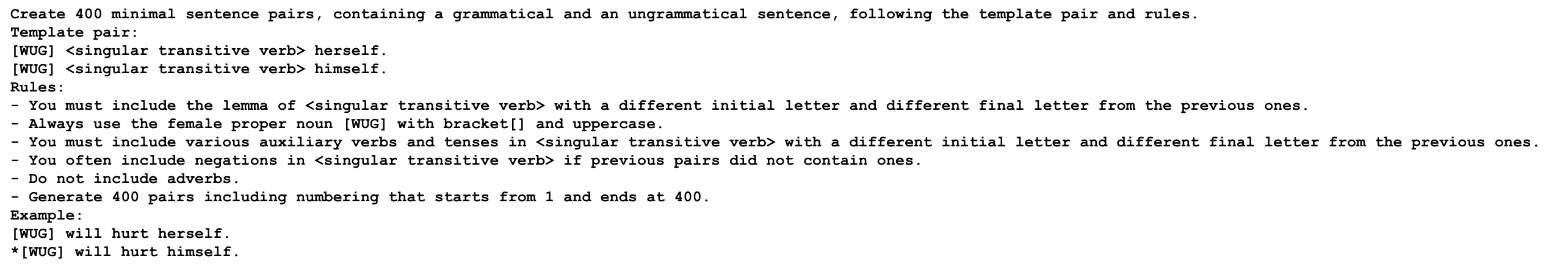}
\caption{An example of prompt used to create evaluation examples.} 
\label{fig:prompt_example}
\end{figure*}

\appendix
\label{sec:appendix}

\section{Data generation}
\label{sec:detailed_data_generation}

\subsection{Linguistic phenomena}
\label{sec:from_blimp}
We employ seven linguistic phenomena, following ~\cite{warstadt-etal-2020-blimp-benchmark}, to create training/evaluation instances.
The linguistic phenomenon ``transitive'' is from ``causative'', ``intransitive'' is from ``drop\_arguement'', ``determiner-noun agreement'' is from ``determiner\_noun\_agreement\_2'', ``subject-verb agreement (V)'' is from ``regular\_plural\_subject\_verb\_agreement\_1'', and ``subject-verb agreement (S)'' is from ``regular\_plural\_subject\_verb\_agreement\_2''.

\subsection{Pretraining Data}
We aim to pretrain the language models for 18 epochs while controlling the number of occurrences of target instances.
To achieve this, we concatenate the pretraining data 18 times consecutively and randomly select where to inject each additional training instance.

\subsection{Creating Data with LLM}
The GPT-4 sometimes inconsistently generates sentences with hallucination; it generates the same sentence repeatedly and sometimes stops generating midway.
To generate as many lexically diverse instances as possible, we prompt GPT-4 to avoid using the same lemma as in the previous instance.
To get appropriate instances, we prompt the GPT-4 to generate double the number of instances\footnote{The number of instances generated based on the prompt can vary. Sometimes the output meets the specified quantity, while other times it may be fewer, potentially even less than half of the requested amount. If not enough instances are generated, we input instances from three steps earlier and generate additional instances to meet the requirements.}, and then select the designated number of instances, avoiding duplicates.
We adjust the percentage of sentences with negation words to 10--50\%.
The balanced instances contained 100 feminine and 100 masculine instances in \textsc{Ana.gen.agr}, 34 feminine singular and 33 masculine singular, 34 singular and 100 plural instances in \textsc{Ana.num.agr}, 200 instances each in \textsc{Transitive} and \textsc{Intransitive}, 50 this, 50 that, 50 these and 50 those in \textsc{D-N agr}. 100 singular and 100 plural each in \textsc{S-V agr}.

\subsection{Prompts}
\label{sec:all_other_prompt_examples}
An example of prompts used to generate minimal sentence pair in anaphor gender agreement where a <wug\#n> in the correct sentence is ``herself'' is shown in Figure~\ref{fig:prompt_example}.
Another example is found in \url{https://github.com/nii-cl/widet}.
We use \texttt{gpt-4-turbo} with top\_p set to 1.0 and temperature set to 0.

\section{Considering BLiMP Score Calculation}
To select one sentence in each pair while evaluating, we calculate its sentence-level likelihood, referring to ~\citet{warstadt-etal-2020-blimp,huebner-etal-2021-babyberta}.
Conversely, \citet{hu-etal-2020-systematic} argue that token-level likelihood comparisons, comparing the aggregate likelihood over a word like "herself" vs. a word like "himself", is a more precise evaluation than sentence-level probability.
We consider the difference using the two phenomena as a case study.

\paragraph{Settings}
We compare the sentence-level likelihood used in this work with two types of score calculation; wug-level likelihood and antecedent-level likelihood.
Given the sentence ``<wug\#n> has devoted herself/*himself,'' the antecedent-level likelihood compares the probabilities assigned to the antecedents ``herself'' and ``himself.'' This is similar to the method used by \citet{hu-etal-2020-systematic}.
The wug-level likelihood, on the other hand, compares the probabilities assigned to each pair of <wug\#n>. 
Since we are using MLMs in our research, it is possible to adapt this for our calculations.

\begin{figure}[tbp]
    \centering
    \includegraphics[width=\linewidth]{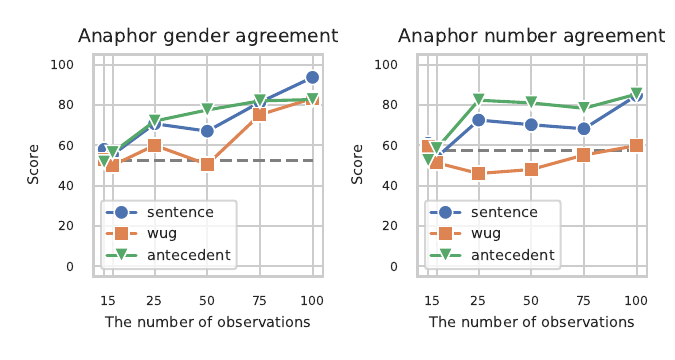}
    \caption{Model's score for three different score calculation methods}
    \label{fig:blimp-cal-level}
\end{figure}

\paragraph{Results}
The score of language models calculated by the different score calculation methods are shown in Figure~\ref{fig:blimp-cal-level}.
Two phenomena are different trends.
For anaphor gender agreement, the sentence-level and wug-level calculation methods show similar trends where the score increased gradually between 25 and 75 occurrences.
The antecedent-level method does not show such a result but hits a plateau after 75 observations.
For anaphor number agreement, the sentence-level and antecedent-level methods show similarities but the latter shows a bit more efficient learning than the former.
The wug-level method does not show improvement until 100 observations.
The results suggest that, in our limited setting, there are distinct trends among the three methods.
The sentence-level and antecedent-level methods each have their advantages depending on the language phenomena.
More analyses of their difference are interesting for future work.

\begin{table}[tbp]
    \centering
    \begin{tabular}{@{}llc@{}}
        \toprule
        \multirow{7}{*}{Model}  
        & architecture & roberta-base \\
        & vocab size & 9,600 \\
        & hidden size & 512 \\
        & heads & 8 \\
        & layers & 8 \\
        & dropout & 0.1 \\
        & layer norm eps & 1e-12 \\
        & initializer range & 0.02 \\
        \cmidrule(r){1-1} \cmidrule(lr){2-2} \cmidrule(lr){3-3}
        \multirow{5}{*}{Optimizer}               
        & algorithm & AdamW \\
        & learning rates & 2e-4 \\
        & betas & (0.9, 0.999) \\
        & weight decay & 0.0 \\
        \cmidrule(r){1-1} \cmidrule(lr){2-2} \cmidrule(lr){3-3}
        \multirow{2}{*}{Scheduler}               
        & type  & linear \\
        & warmup updates & 24,000 \\
        \cmidrule(r){1-1} \cmidrule(lr){2-2} \cmidrule(lr){3-3}
        \multirow{3}{*}{Training}                
        & gradient accum. & 4 \\
        & epoch & 18 \\
        & batch size & 16 \\
        & line by line & true \\
        & NGPU & 1 \\
        \bottomrule
    \end{tabular}
\caption{Hyperparameters of the language models.}
\label{tab:hypara}
\end{table}

\section{Hyperparameters}
\label{sec:hypara}

Hyperparameters in our work are listed in Table~\ref{tab:hypara}.

\end{document}